\documentclass[runningheads]{llncs}

\usepackage{eccv}

\usepackage{eccvabbrv}

\usepackage{graphicx}
\usepackage{booktabs}
\usepackage{bbding}
\usepackage{algpseudocode}
\usepackage{algorithm}
\usepackage[accsupp]{axessibility}  

\usepackage{hyperref}

\usepackage{orcidlink}
\usepackage{multirow}

\begin{document}

\title{AccDiffusion: An Accurate Method for Higher-Resolution Image Generation} 

\titlerunning{An Accurate Method for Higher-Resolution Image Generation}


\author{Zhihang Lin\inst{1} \and
Mingbao Lin\inst{2} \and
Meng Zhao \inst{3} \and Rongrong Ji \inst{1}\thanks{Corresponding author} 
}

\authorrunning{Z.Lin et al.}

\institute{Key Laboratory of Multimedia Trusted Perception and Efficient Computing, Ministry of Education of China, Xiamen University, China. \and Skywork AI. \and Tencent Youtu Lab. \\
\email{zhihanglin@stu.xmu.edu.cn,} \email{linmb001@outlook.com,}
\email{arthurrizar8421@gmail.com,}
\email{rrji@xmu.edu.cn} 
{\small \url{https://lzhxmu.github.io/accdiffusion/accdiffusion.html}}
\vspace{-1em}
}

\maketitle

\begin{abstract}
%

This paper attempts to address the object repetition issue in patch-wise higher-resolution image generation. We propose AccDiffusion, an accurate method for patch-wise higher-resolution image generation without training. An in-depth analysis in this paper reveals an identical text prompt for different patches causes repeated object generation, while no prompt compromises the image details. Therefore, our AccDiffusion, for the first time, proposes to decouple the vanilla image-content-aware prompt into a set of patch-content-aware prompts, each of which serves as a more precise description of an image patch. Besides, AccDiffusion also introduces dilated sampling with window interaction for better global consistency in higher-resolution image generation. Experimental comparison with existing methods demonstrates that our AccDiffusion effectively addresses the issue of repeated object generation and leads to better performance in higher-resolution image generation.

\end{abstract}

\section{Introduction}
\label{sec:intro}

Diffusion models have garnered significant attention and made notable advancements with the emergence of works such as DDPM~\cite{ho2020ddpm}, DDIM~\cite{song2020ddim}, ADM~\cite{dhariwal2021ldm}, and LDMs~\cite{rombach2022SD}, owing to their outstanding generative ability and wide range of applications. 
However, stable diffusion models entail tremendous training costs primarily due to the large number of timestamps required and the quadratic relationship between computing costs and resolution.
Consequently, it is common to limit the resolution to a relatively low level, such as $512^2$ for SD $1.5$~\cite{stable-diffusion-1.5} and $1024^2$ for SDXL~\cite{podell2023sdxl}, during training. 
Even at such low resolution, stable diffusion 1.5 still entails over 20 days of training on 256 A100 GPUs~\cite{stable-diffusion-1.5}. 
Nonetheless, high-resolution generation finds widespread application in real-life scenarios, such as advertisements. The demand for generating high-resolution images clashes with the expensive training costs involved.

\begin{figure}[!t]
    \centering
    \includegraphics[width=\linewidth]{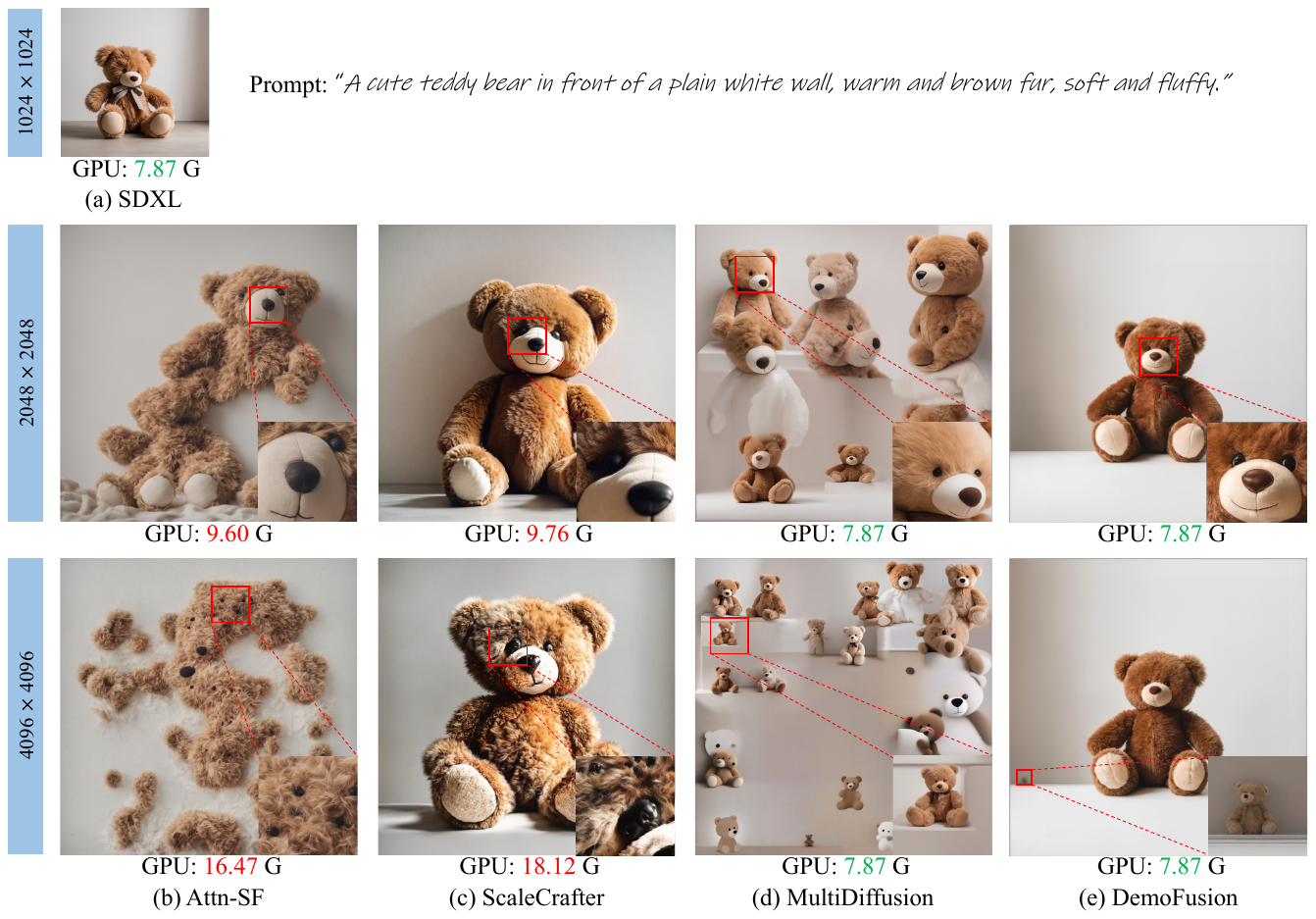}
    \caption{Comparison of image quality and GPU overhead for existing higher-resolution generation methods. The GPU memory of  Attn-SF~\cite{jin2023logn} and ScaleCrafter~\cite{he2023scalecrafter} significantly increases with resolution, while patch-wise denoising methods ,\eg, MultiDiffusion~\cite{bar2023multidiffusion} and DemoFusion~\cite{du2023demofusion} suffer object repetition issue. Best viewed zoomed in. 
    }
    \label{fig:Related work Comparison}
\end{figure}

Therefore, researchers have shifted their focus to training  stable diffusion models with low resolution and subsequently applying fine-tuning~\cite{zheng2023any-size-diffusion,xie2023diff-fit} or training-free~\cite{bar2023multidiffusion,du2023demofusion,he2023scalecrafter,lee2023syncdiffusion} methods to achieve image generation extrapolation.
A naive approach is to directly use pre-trained stable diffusion models to generate higher-resolution images. However, the resulting images from this approach are proved to suffer from issues such as object repetition and inconsistent object structures~\cite{jin2023logn,du2023demofusion}. Previous methods attempted to achieve image generation extrapolation from the perspectives of attention entropy~\cite{jin2023logn} or the receptive field of stable diffusion model~\cite{he2023scalecrafter}. However, these methods have been proven to be less practical in two folds, as shown in Fig.\,\ref{fig:Related work Comparison}(b,c): 
(1)a substantial increase in GPU memory consumption~\cite{zheng2023any-size-diffusion} as the resolution rises and (2) poor quality of the generated images~\cite{du2023demofusion}.
Thanks to stable diffusion's outstanding local detail generation ability, recent works~\cite{du2023demofusion,lee2023syncdiffusion,bar2023multidiffusion} have started conducting higher-resolution image generation in a patch-wise fashion for the sake of less GPU memory consumption.
Previous works MultiDiffusion~\cite{bar2023multidiffusion} and SyncDiffusion~\cite{lee2023syncdiffusion} fuse multiple overlapped patch-wise denoising results to generate higher-resolution panoramic images without a seam. 
However, the direct application of these approaches to generate higher-resolution object-centric images leads to repeated and distorted results lacking global semantic coherence, as shown in Fig.\,\ref{fig:Related work Comparison}(d).
Recently, DemoFusion~\cite{du2023demofusion} has introduced global semantic information into the patch-wise higher-resolution image generation through  residual connection and dilated sampling.
It only partially solves the problem of repeated object generation 
and still exhibits small object repetition in ultra-high 
image generation as depicted in Fig.\,\ref{fig:Related work Comparison}(e).
How to resolve the issue of repeated object generation completely in patch-wise higher-resolution image generation remains an unresolved problem.

\begin{figure}[tb]
    \centering
    \includegraphics[width=\linewidth]
    {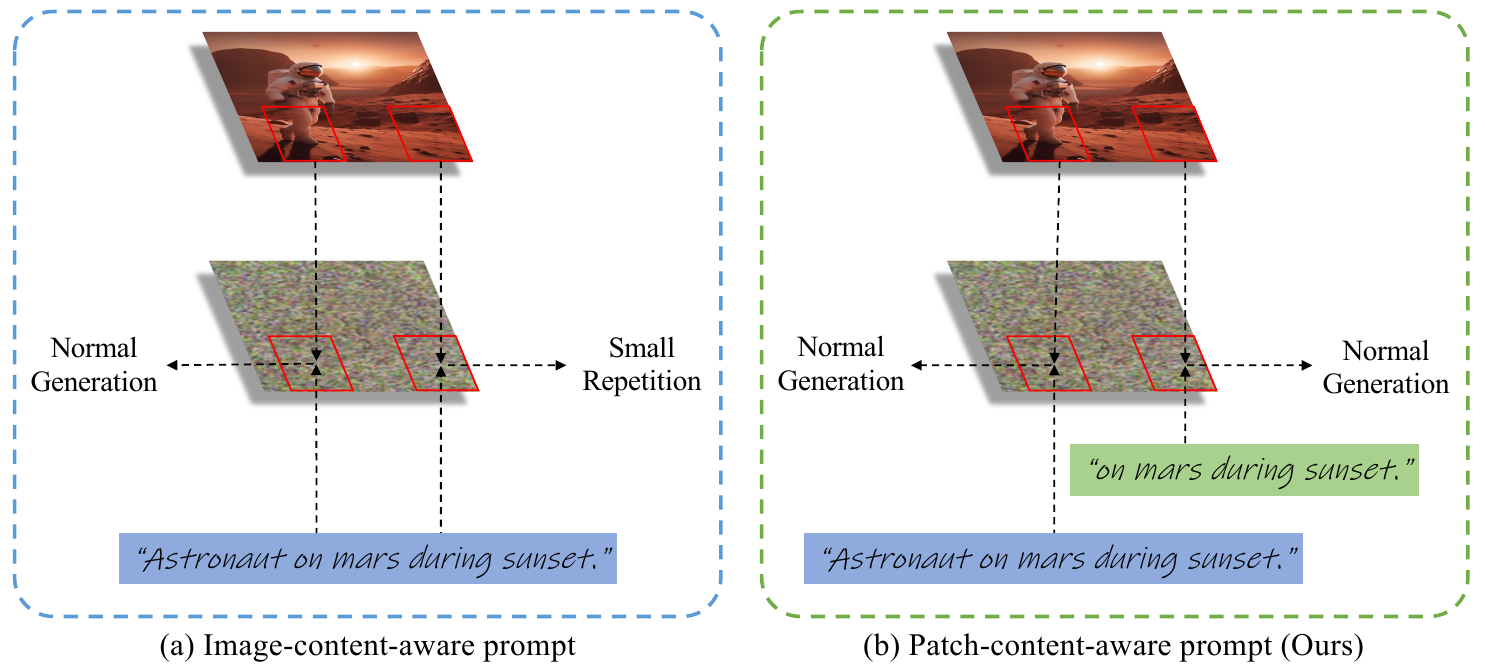}
    \caption{Image-content-aware prompt \emph{v.s.} Patch-content-aware prompt.}
    \label{fig:repetition analyze}
\end{figure}


In this paper, our in-depth analysis of DemoFusion~\cite{du2023demofusion} indicates, as illustrated in Fig.\,\ref{fig:repetition analyze}(a), small object repetition generation is the adversarial outcome of an identical text prompt on all patches, encouraging to generate repeated objects, and global semantic information from residual connection and dilated sampling, suppressing the generation of repeated objects.
To address the above issues, we propose AccDiffusion, an accurate method for higher-resolution image generation, with its major novelty in two folds:

(1) To completely solve small object repetition, as illustrated in  Fig.\,\ref{fig:repetition analyze}(b), we propose to decouple the vanilla image-content-aware prompt into a set of patch-content-aware substrings, each of which serves as a more precise prompt to describe the patch contents. Specifically, we utilize the cross-attention map from the low-resolution generation process to determine whether a word token should serve as the prompt for a patch.
If a word token has a high response in the cross-attention map region corresponding to the patch, it should be included in the prompt, and vice versa.

(2) Through visualization, we observe that the dilated sampling operation in DemoFusion generates globally inconsistent and noisy information, disrupting the generation of higher-resolution images. 
Such inconsistency stems from the independent denoising of dilation samples without interaction. 
To address this, we employ a position-wise bijection function to enable interaction between the noise from different dilation samples. 
Experimental results show that our dilated sampling with interaction leads to the generation of smoother global semantic information (see Fig.\,\ref{fig:ablation analyze}(c,d)).

We have conducted extensive experiments to verify the effectiveness of AccDiffusion.
The qualitative results demonstrate that AccDiffusion effectively addresses the issue of repeated object generation in higher-resolution image generation.
And the quantitative results show that AccDiffusion achieves state-of-the-art performance in training-free image generation extrapolation.



\section{Related Work}

\subsection{Diffusion Models}
Diffusion models~\cite{ho2020ddpm,song2020ddim,dhariwal2021ldm,rombach2022SD} are generative probabilistic models that transform Gaussian noise into samples through gradual denoising steps. 
DDPM~\cite{ho2020ddpm} is a pioneering model that demonstrates impressive image generation capabilities using Markovian forward and reverse processes.
Based on DDPM, DDIM~\cite{song2020ddim} utilizes non-Markovian reverse processes to decrease sampling time effectively. 
Furthermore, LDMs~\cite{rombach2022SD} incorporate the diffusion process into the latent space, resulting in efficient training and inference.
Subsequently, a series of LDMs-based stable diffusion models are open-sourced and achieve state-of-the-art image synthesis capability.
This has led to widespread applications in various downstream generative tasks, including images~\cite{dhariwal2021ldm,ho2020ddpm,nichol2021improved,song2020ddim,saharia2022photorealistic}, audio~\cite{huang2023make-a-audio,ghosal2023Text-to-Audio}, video~\cite{singer2022make-a-video,ho2022imagen} and 3D objects~\cite{lin2023magic3d,xu2023dream3d,poole20223d-dreamfusion}, \emph{etc}.

\subsection{Training-Free Higher-Resolution Image Generation} 
Although stable diffusion demonstrates impressive results, its training cost limits low-resolution training and thus generates low-fidelity images when the inference resolution differs from the training resolution~\cite{jin2023logn,he2023scalecrafter,du2023demofusion}.
%
Recent works~\cite{jin2023logn,he2023scalecrafter,du2023demofusion,bar2023multidiffusion} have attempted to utilize pre-trained diffusion models for generating higher-resolution images.
These works~\cite{jin2023logn,he2023scalecrafter,du2023demofusion,bar2023multidiffusion}  can be broadly categorized into two categories: direct generation~\cite{he2023scalecrafter,jin2023logn} and indirect generation~\cite{bar2023multidiffusion,du2023demofusion}.
Direct generation methods scale the input of the diffusion models to the target resolution and then perform forward and reverse processes directly on the target resolution.
These kinds of methods require modifications to the fundamental architecture, such as adjusting the attention scale factor~\cite{jin2023logn} and the receptive field of convolutional kernels~\cite{he2023scalecrafter}, to prevent repetition generation.
However, the generated images fail to yield the higher-resolution detail desired.
Additionally, direct generation methods encounter out-of-memory errors when generating ultra-high resolution images (\eg 8K) on consumer-grade GPUs, due to the quadratic increase in memory overhead as the latent space size grows.
Indirect generation methods generate higher-resolution images through multiple overlapped denoising paths of LDMs and are capable of generating images of any resolution on consumer-grade GPUs. 
However, these mothods~\cite{bar2023multidiffusion,lee2023syncdiffusion} suffer from local repetition and structural distortion.
Du \etal~\cite{du2023demofusion} tried to address repeated generation by introducing global structural information from lower-resolution image. 
%

\section{Method}

\subsection{Backgrounds}\label{sec:backgrounds}

\textbf{Latent Diffusion Models (LDMs)}.
LDMs~\cite{dhariwal2021ldm} apply an autoencoder $\mathcal{E}$ to encode an image $\mathbf{x}_0 \in \mathbb{R}^{H\times W \times 3}$ into a latent representation $\mathbf{z}_0 = \mathcal{E}(\mathbf{x}_0) \in \mathbb{R}^{h\times w \times c}$, where the regular diffusion process is constructed as:
\begin{equation}
    \mathbf{z}_t = \sqrt{\bar{\alpha}_t}\mathbf{z}_0 + \sqrt{1-\bar{\alpha}_t}\varepsilon,\quad \varepsilon\sim \mathcal{N}(0,\mathbf{I}),
\end{equation}
where $\{{\alpha}_t\}_{t=1}^T$  is a set of prescribed variance schedules and $\bar{\alpha}_t = \Pi_{i=1}^t \alpha_i$. 
%
%
%
%
%
To perform conditional sequential denoising, a network $\varepsilon_{\theta}$ is trained to predict added noise, constrained by the following training objective:
\begin{equation}
\label{SD_training_loss}
    \underset{\theta}{\text{min}}\ \mathbb{E}_{\mathcal{E}(x_0), \varepsilon \sim \mathcal{N}(0,1), t\sim \text{Uniform}(1, T)}\Big[\left\| \varepsilon - \varepsilon_{\theta}\big(\mathbf{z}_t, t, \tau_{\theta}(y)\big)\right\|_2^2\Big],
\end{equation}
%
in which $\tau_{\theta}(y)\in \mathbb{R}^{M\times d_{\tau}}$ is an intermediate representation of condition $y$ and $M$ is the number of word tokens in the prompt $y$. The $\tau_{\theta}(y)$ is then mapped to keys and values in cross-attention of U-Net $\varepsilon_{\theta}$: 

\begin{equation}
\label{cross-attention}
\begin{aligned}
        Q = W_Q\cdot \varphi(z_t),\quad K = W_K\cdot \tau_{\theta}(y),\quad V = W_V\cdot \tau_{\theta}(y), \\  
    \mathcal{M} = \text{Softmax}(\frac{QK^T}{\sqrt{d}}),\quad \text{Attention}(Q,K,V) = \mathcal{M} \cdot V.
\end{aligned}
\end{equation}

Here, for simplicity, we omit the expression of multi-head cross-attention and $\varphi(z_t)\in \mathbb{R}^{N\times d_\epsilon}$ denotes an intermediate representation of noise in the U-Net. 
Here $N = h \times w$ represents the pixel number of the latent noise $z_t$.
$W_Q \in \mathbb{R}^{d\times d_{\epsilon}}, W_K \in \mathbb{R}^{d\times d_{\tau}} $, and $
 W_V\in \mathbb{R}^{d\times d_{\tau}} $ are learnable projection matrices. $\mathcal{M} \in \mathbb{R}^{N\times M}$ is the cross-attention maps.

In contrast, the denoising process aims to recover the cleaner version $\mathbf{z}_{t-1}$ from $\mathbf{z}_{t}$ by estimating the noise, which can be expressed as:
\begin{equation}
\label{denoising process}
    \mathbf{z}_{t-1} = \sqrt{\frac{\alpha_{t-1}}{\alpha_t}}\mathbf{z}_t + \Bigg(\sqrt{\frac{1}{\alpha_{t-1}} - 1} - \sqrt{\frac{1}{\alpha_t} - 1} \Bigg)\cdot \varepsilon_{\theta}\big(\mathbf{z}_t,t,\tau_{\theta}(y)\big).
\end{equation}

During inference, a decoder $\mathcal{D}$ is employed at the end of the denoising process to reconstruct the image from the latent representation $\mathbf{x}_0 = \mathcal{D}(\mathbf{z}_0)$.

\textbf{Patch-wise Denoising}.
MultiDiffusion~\cite{bar2023multidiffusion} achieve higher-resolution image generation by fusing multiple overlapped denoising patches.
In simple terms, given a latent representation $\mathcal{Z}_t \in \mathbb{R}^{h'\times w'\times c}$ of higher-resolution image with $h' > h$ and $w' > h$, MultiDiffusion utilizes a shifted window to sample patches from $\mathcal{Z}_t$ and results in a series of patch noise $\{\mathbf{z}_t^i\}_{i=1}^{P_1}$, where $\mathbf{z}_t^i \in \mathbb{R}^{h\times w\times c}$ and $P_1 = (\frac{h'-h}{d_h}+1)\times(\frac{w'-w}{d_w}+1)$ is the total number of patches, $d_h$ and $d_w$ is the vertical and horizontal stride, respectively.
Then, MultiDiffusion performs patch-wise denoising via Eq.\,(\ref{denoising process}) and obtains $\{\mathbf{z}_{t-1}^i\}_{i=1}^{P_1}$.
Then $\{\mathbf{z}_{t-1}^i\}_{i=1}^{P_1}$ is reconstructed to get $\mathcal{Z}_{t-1}$, where the overlapped parts take the average. Eventually, a higher-resolution image can be obtained by directly decoding  $\mathcal{Z}_0$ into image  $\mathbf{X}_0$.
Based on MultiDiffusion, DemoFusion~\cite{du2023demofusion} additionally introduces:
1) progressive upscaling to gradually generate higher-resolution images;
2) residual connection to maintain global consistency with the lower-resolution image by injecting the intermediate noise-inversed representation.
3) dilated sampling to enhance global semantic information of higher-resolution images.


\begin{figure}[!t]
    \centering
    \includegraphics[width=\linewidth]{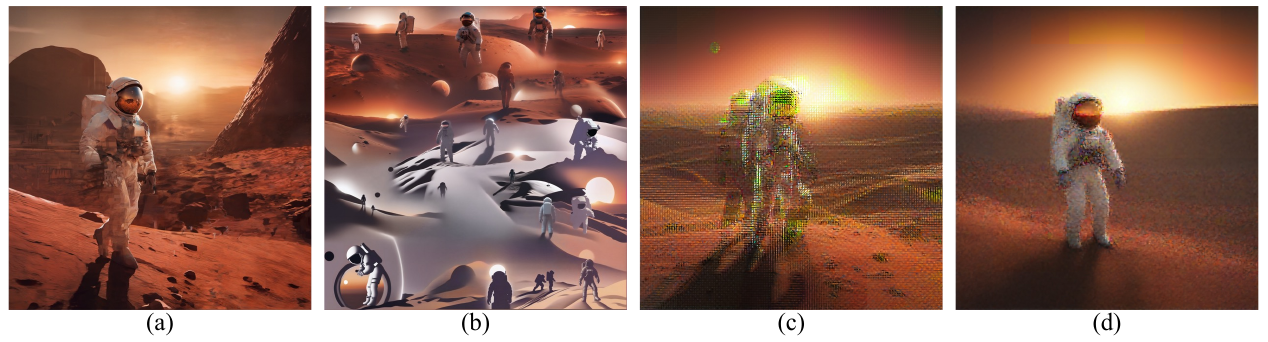}
    \caption{Results of higher-resolution image generation. (a) The result of DemoFusion without text prompt. (b)The result of DemoFusion without residual connection and dilated sampling. (c) The result of dilated sampling without window interaction. (d)The result of our dilated sampling with window interaction.}
    \label{fig:ablation analyze}
\end{figure}

\subsection{In-depth Analysis of Small Object Repetition}

DemoFusion demonstrates the possibility of using pre-trained LDMs to generate higher-resolution images.
However, as shown in Fig.\,\ref{fig:Related work Comparison}(e), small object repetition continues to challenge the performance of DemoFusion.

%

Delving into an in-depth analysis, we respectively:
1) remove the text prompt during higher-resolution generation of DemoFusion and the resulting Fig.\,\ref{fig:ablation analyze}(a) indicates the disappearance of repeated objects but more degradation in details.
2) remove the operations of residual connection \& dilated sampling in DemoFusion and the resulting Fig.\,\ref{fig:ablation analyze}(b) denotes severe large object repetition.
Therefore, we can make a safe conclusion that  small object repetition is the adversarial outcome of an identical text prompt on all patches and operations of residual connection \& dilated sampling.
The former encourages to generate repeated objects while the latter suppresses the generation of repeated objects. 
Consequently, DemoFusion tends to generate small repeated objects.


Overall, text prompts play a significant role in image generation. It is not a viable solution to address small object repetition by removing text prompts during the higher-resolution generation, as it would lead to a decline in image quality. 
Instead, we require more accurate prompts specifically tailored for each patch.
That is, if an object is not present in a patch, the corresponding word in the text prompts should not serve as a prompt for that patch.

To this end, in Sec.\,\ref{patch-level prompts}, we eliminate the restriction of having an identical text prompt for all patches in previous patch-wise generation approaches. Instead, we generate more precise patch-content-aware prompts that adapt to the content of different patches. 
In Sec.\,\ref{Shuffle Dilated-Sampling}, we introduce how to enhance the global structure information to generate higher-resolution images without repetition.

\subsection{Patch-Content-Aware Prompts}
\label{patch-level prompts}

\begin{figure}[tb]
    \centering
    \includegraphics[width=\linewidth]{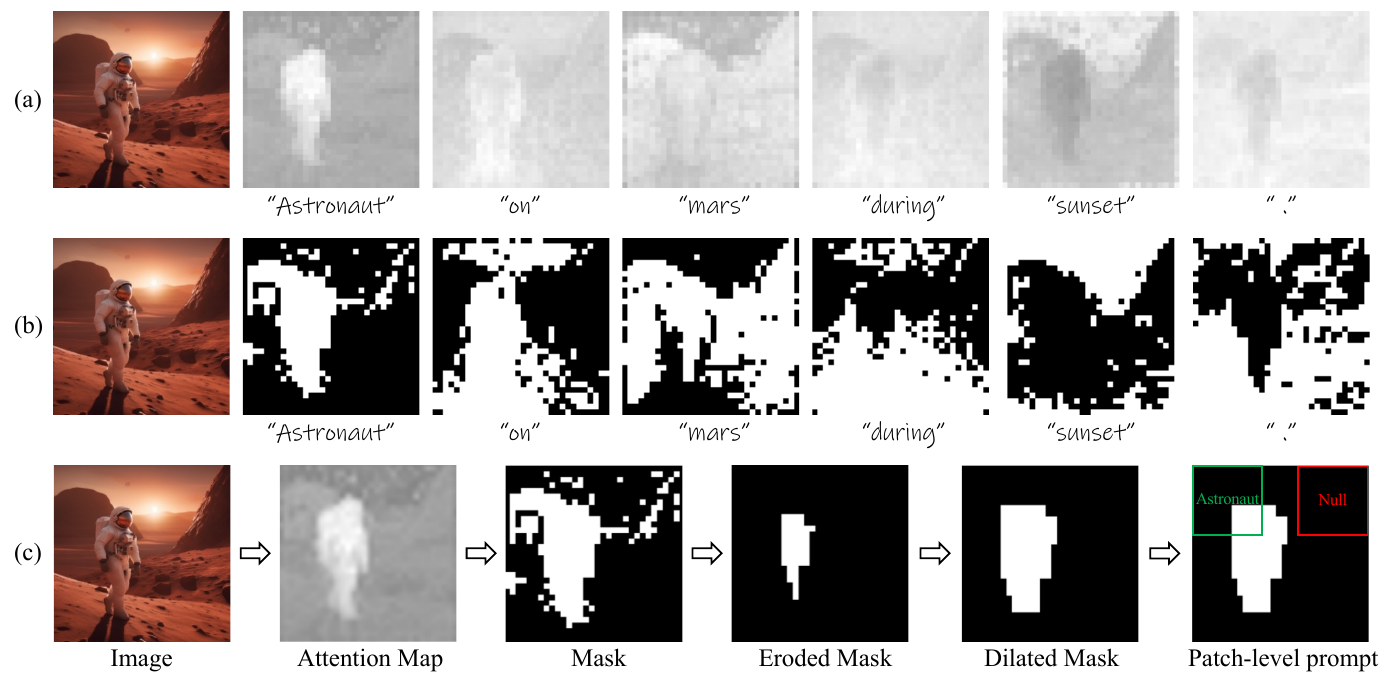}
    \caption{
    Visualization of averaged attention map from the up blocks and down blocks in U-Net. We reshape the attention map into a 2D shape before visualization.
    (a) Cross-attention map visualization using open source code~\cite{hertz2022prompt-to-prompt}.  
    (b) Highly responsive regions of each word. 
    (c) The illustration of the patch-level prompt generation process, including morphological operations to eliminate small connected areas. Here we use the word ``Astronaut'' as an example. All words in the prompt will go through the above process.}
    \label{fig:attention_map}
\end{figure}

Considering the significance of text prompt in higher-resolution generation, we explore patch-content-aware substring set $\{ \gamma^i\}_{i=1}^{P_1}$ of the entire text prompt, each of which is responsible for injecting a condition to the corresponding patch.
In general, it is challenging to know in advance what content a patch generates, but in DemoFusion~\cite{du2023demofusion}, the global information from low-resolution image is injected into the high-resolution image generation through residual connections. 
Therefore, the structure of the generated higher-resolution image is similar to that of the low-resolution image.
%
%
This inspires us to decide patch contents from the low-resolution image.
A direct but cumbersome approach is to manually observe the patch content of low-resolution image and then set the prompt for each patch, which undermines the usability of stable diffusion.
Another approach is to use SAM~\cite{kirillov2023sam} to segment the upscaled low-resolution image and determine whether each object appears in the patch, introducing huge storage and computational costs of the segmentation model.
How to automatically generate patch-content-aware prompts without external models is the key to success.

Inspired by image editing~\cite{hertz2022prompt-to-prompt}, instead we consider the cross-attention maps in low-resolution generation $\mathcal{M} \in \mathbb{R}^{N \times M}$, to determine patch-content-aware prompts.
Recall $N$ represents the pixel number of the latent noise $z_t$ and $M$ denotes the number of word tokens in the prompt $y$.
Thus, the column $\mathcal{M}_{:, j}$ represents the attentiveness of latent noise to the $j$-th word token.
The basic principle lies in that the attentiveness ($\mathcal{M}_{i, j}$) of image regions is mostly higher than others if it is attended by the $j$-th word token, as shown in Fig.\,\ref{fig:attention_map}(a). 
To find the highly relevant region of each word token, we convert the attention map $\mathcal{M}$ into a binary mask $\mathcal{B} \in \mathbb{R}^{N \times M}$ as:
\begin{equation}\label{eq:highly responsive regions}
\mathcal{B}_{i,j} = \left\{
\begin{array}{ll}
1 , \textrm{\; if $\mathcal{M}_{i,j}$ > $\overline{\mathcal{M}}_{:,j}$},\\
0 , \textrm{\; otherwise},
\end{array}\right.
\end{equation}
where $i$ and $j$ enumerate $N$ and $M$, respectively. 
%
The threshold $\overline{\mathcal{M}}_{:,j}$ is the mean of $\mathcal{M}_{:,j}$, which design is
elaborated in Sec.\,\ref{ablaiton study}.
Regions with values above the threshold are considered highly responsive, while regions with values below the threshold are considered less responsive.

Next, we obtain word-level masks $\{\mathcal{B}_j\}^M_{j=1}$ using the following equation:
\begin{equation}
\hat{\mathcal{B}}_j = \text{Reshape}(\mathcal{B}_{:,j},(h_a,w_a)), 
\end{equation}
where $h_a = \frac{h}{s}$ and $w_a = \frac{w}{s}$ represent the height and width of the attention map, respectively. 
Recall $h$ and $w$ represent the height and width of the noise, respectively.
The ``$s$'' corresponds to the down-sampling scale in the corresponding block of the U-Net model. The mask ${\mathcal{B}}_j$ oriented for the $j$-th word token is reshaped into a 2d shape for further processing.

After obtaining the highly responsive regions for each word, we observe that they contain many small connected areas, as shown in Fig.\,\ref{fig:attention_map}(b).
%
%
To alleviate the influence of these small connected areas, we apply the opening operation $\mathcal{O}(\cdot)$ from mathematical morphology~\cite{soille1999morphological}, resulting in the final mask for each word, as shown in Fig.\,\ref{fig:attention_map}(c).
The processed mask $\{\Tilde{\mathcal{B}}_j\}^M_{j=1}$ can be formulated as:
\begin{equation}
 \Tilde{\mathcal{B}}_j = \mathcal{O}(\hat{\mathcal{B}}_j) = \omega(\delta(\hat{\mathcal{B}}_j)),
\end{equation}
where $\delta(\cdot)$ and $\omega(\cdot)$ is erosion operation  and  dilation operation, respectively.
Next, we interpolate $ \Tilde{\mathcal{B}}_j \in \mathbb{R}^{h_a\times w_a}$ to $ \Tilde{\mathcal{B}}'_j \in \mathbb{R}^{h'_a\times w'_a}$, where $h'_a = \frac{h'}{s}$ and $w'_a = \frac{w'}{s}$.
Recall $h'$ and $w'$ are the size of higher-resolution latent representation as defined in Sec.\,\ref{sec:backgrounds}.
Inspired by MultiDiffusion~\cite{bar2023multidiffusion}, we use a shifted window to sample patches from $\Tilde{\mathcal{B}}'_j$ , resulting in a series of patch masks $\{\{\mathbf{m}^i_j\}_{i=1}^{P_1}\}_{j=1}^{M}$, where $\mathbf{m}^i_j \in \mathbb{R}^{h_a\times w_a}$ and $P_1$ is the total number of patches. 
It is important to note that each $\mathbf{m}^j_i$ corresponds to a specific patch noise $\mathbf{z}_t^i$.

Recall if an object is not present in a patch, the corresponding word token in the text prompts should not serve as a prompt for that patch.
So, we can determine the patch-content-aware prompt $\gamma^i$, a sub-sequence of prompt $y$, for each patch $\mathbf{z}_t^i$ using the following formulation:
\begin{equation}
\label{eq:patch-content-aware prompt}
\left\{
\begin{array}{ll}
y_j \in \gamma^i, \textrm{\; if $\frac{\sum(\mathbf{m}^i_j)_{:,:}}{h_a \times w_a}$ > $c$},\\
y_j \notin \gamma^i, \textrm{\; otherwise},
\end{array}\right.
\end{equation}
where $j$ and $i$ enumerates $M$ and $P_1$, respectively. The pre-given hyper-parameter $c\in (0,1)$ determines whether a highly responsive region's proportion of a word $y_j$ exceeds the threshold for inclusion in the prompts of patch $z_t^i$.
We then concatenate all words that should appear in a patch together, resulting in patch-content-aware prompts $\{\gamma^i\}_{i=1}^{P_1}$ for noise patches $\{z_t^i\}_{i=1}^{P_1}$ during patch-wise denoising.

\begin{figure}[tb]
    \centering
    \includegraphics[width=\linewidth]{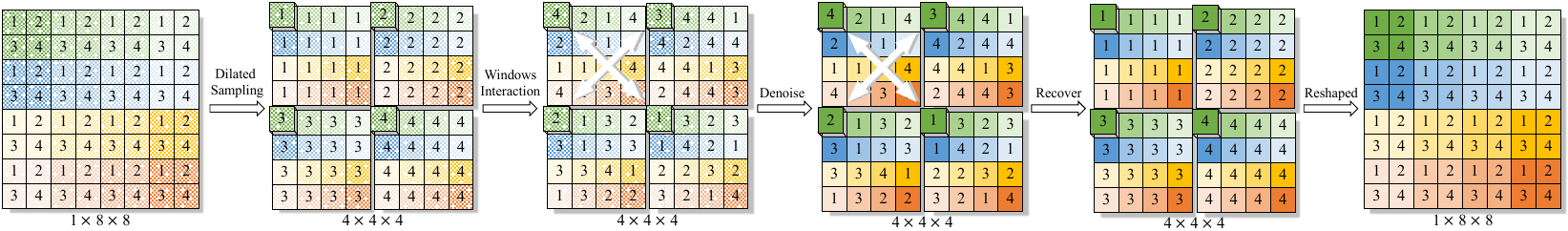}
    \caption{Illustration of dilated sampling with window interaction: $8 \times 8$ higher-resolution and $4 \times 4$ low-resolution. 
    The number $\{1,2,3,4\}$ represent the different positions within the same window (same color). The interaction operation is conducted in the window.}
    
    \label{fig:dilated sampling pipline}
\end{figure}

\subsection{ Dilated Sampling with Window Interaction}
\label{Shuffle Dilated-Sampling}

Recall $\mathcal{Z}_t \in \mathbb{R}^{h'\times w'\times c}$ stands for the latent representation of a higher-resolution image in Sec.\,\ref{sec:backgrounds}.
In this section, we continue proposing dilated sampling with window interaction, for a set of patch samples $\{D_t^k\}_{k=1}^{P_2}$, to improve the global semantic information in the latent representation $\mathcal{Z}_t$. 
In DemoFusion~\cite{du2023demofusion}, each sample $D_t^k$ is a subset of the latent representation $\mathcal{Z}_t$, formulated as:
\begin{equation}
    \mathcal{D}_t^{k} = (\mathcal{Z}_{t})_{i::h_s,j::w_s,:},
\end{equation}
where $k = i \times w_s + j + 1$, and $k$ ranges from $1$ to $P_2$. The variables $i$ and $j$ range from $0$ to $h_s-1$ and $w_s-1$, respectively.
The sampling stride is determined by $h_s = \frac{h'}{h}$ and $w_s= \frac{w'}{w}$. Recall $\{h',w'\}$ and $\{h, w\}$ are the height and width of  higher and low resolution latent representation.
DemoFusion independently performs denoising on $\mathcal{D}_t$ via Eq.\,(\ref{denoising process}) and obtains $\mathcal{D}_{t-1} \in \mathbb{R}^{P_2 \times  h \times w \times c}$, where $P_2 = h_s \times w_s$.
Then $\{\mathcal{D}_{t-1}^k\}_{k=1}^{P_2}$ is reconstructed as $G_{t-1} \in \mathbb{R}^{h'\times w' \times c}$ and added to patch-wise denoised latent representation ${\mathcal{Z}_{t-1}}$ using:
\begin{equation}
\label{dilated sampling }
\begin{aligned}
    {\hat{\mathcal{Z}}_{t-1}} = (1 - \eta) &\cdot {\mathcal{Z}_{t-1}} + \eta \cdot {G_{t-1}}, 
\end{aligned}
\end{equation}
where $(G_{t-1})_{i::h_s,j::w_s,:}  = {\mathcal{D}^{k}_{t-1}}$ and $\eta$ decreases from $1$ to $0$ using a cosine schedule.
Due to the lack of interaction between different samples during the denoising process, the global semantic information is non-smooth, as depicted in Fig.\,\ref{fig:ablation analyze}(c). 
The sharp global semantic information disturbs the higher-resolution generation. 

To solve above issue, as illustrated in Fig.\,\ref{fig:dilated sampling pipline}, we enable window interaction between different samples before each denoising process through bijective function:
\begin{equation}
    {\mathcal{D}_t}^{k,h,w} = {\mathcal{D}_t}^{f_t^{h,w}(k),h,w},\quad f^{h,w}_t:\{1,2,\cdots,P_2\} \Rightarrow \{1,2,\cdots,P_2\},
\end{equation}
where $f^{h,w}_t$ is  bijective function, and the mapping varies based on the position or time step. 
We then perform normal denoising progress on $ \{\mathcal{D}_{t}^k\}_{k=1}^{P_2}$  to obtain $\{\mathcal{D}_{t-1}^k\}_{k=1}^{P_2}$.
Before applying Eq.\,(\ref{dilated sampling }) to $\{\mathcal{D}_{t-1}^k\}_{k=1}^{P_2}$, we use the inverse mapping ${(f^{h,w}_t)}^{-1}$  of $f^{h,w}_t$ to recover the position as:
\begin{equation}
    {\mathcal{D}_{t-1}}^{k,h,w} = {\mathcal{D}_{t-1}}^{{(f^{h,w}_t)}^{-1}(k),h,w}, \quad {(f^{h,w}_t)}^{-1}:\{1,2,\cdots,P_2\} \Rightarrow \{1,2,\cdots,P_2\}.
\end{equation}


\section{Experimentation}

\subsection{Experimental Setup}
AccDiffusion is a plug-and-play extension to stable diffusion without additional training costs.  
We mainly validate the feasibility of AccDiffusion using the pre-trained SDXL~\cite{podell2023sdxl}. 
More results for other stable diffusion variants are in Appendix\,\ref{more variants}.
AccDiffusion follows the pipeline of DemoFusion~\cite{du2023demofusion} (SOTA higher-resolution generation)
and uses the patch-content-aware prompts during the progress of higher-resolution image generation.
Additionally, AccDiffusion enhances dilated sampling with window interaction.
For fairness, we adhere to the default setting of DemoFusion, as described in Appendix\,\ref{app:default setting}. 
In Sec.\,\ref{quantitative comparison}, the hyper-parameter $c$ in Eq.\,(\ref{eq:patch-content-aware prompt}) is set to 0.3.
Considering the training-free nature of  AccDiffusion, the methods we compare include:
\textbf{SDXL-DI}~\cite{podell2023sdxl},
\textbf{Attn-SF}~\cite{jin2023logn},
\textbf{ScaleCrafter}~\cite{he2023scalecrafter},
\textbf{MultiDiffusion}~\cite{bar2023multidiffusion},
and \textbf{DemoFusion}~\cite{du2023demofusion}.
We do not compare with image super-resolution methods~\cite{wang2023sr,saharia2022sr, zhang2021BSRGAN} which take images as input and have been proven to lack texture details~\cite{he2023scalecrafter,du2023demofusion}.


\subsection{Quantitative Comparison}
\label{quantitative comparison}
For quantitative comparison, we use three widely-recognized metrics: FID (Frechet Inception Distance)~\cite{heusel2017fid}, IS (Inception Score)~\cite{salimans2016is}, and CLIP Score~\cite{radford2021clip-score}. Specifically, $\text{FID}_r$ measures the Frechet Inception Distance between generated high-resolution images and real images.
$\text{IS}_r$ represents the Inception Score of generated high-resolution images.
Given that $\text{FID}_r$ and $\text{IS}_r$ necessitate resizing images to $299^2$, which may not be ideal for assessing high-resolution images.
Motivated by~\cite{du2023demofusion,chai2022Any-resolution-training}, we crop 10 local patches at $1\times$ resolution from each generated high-resolution image and subsequently resize them to calculate $\text{FID}_c$ and $\text{IS}_c$.
The CLIP score measures the cosine similarity between image embedding and text prompts.
We randomly selected 10,000 images from the Laion-5B~\cite{schuhmann2022laion-5b} dataset as our real images set and randomly chose 1,000  text prompts from Laion-5B as inputs for AccDiffusion to generate a set of high-resolution images.

As shown in Table\,\ref{tab:quantitative comparison}, AccDiffusion achieves the best results and obtains state-of-the-art performance.
Since the implementation of  AccDiffusion is based on DemoFusion~\cite{du2023demofusion}, it exhibits similar quantitative results and inference times with DemoFusion.
However, AccDiffusion outperforms DemoFusion due to its more precise patch-content-aware prompt and more accurate global information introduced by dilated sampling with interaction, especially in high-resolution generation scenarios.
Compared to other training-free image generation extrapolation methods, the quantitative results of AccDiffusion are closer to quantitative results calculated at pre-trained resolutions (1024 $\times$ 1024), demonstrating the excellent image generation extrapolation capabilities of AccDiffusion.
Note that $\text{FID}$, $\text{IS}$, and CLIP-Score do not intuitively reflect the degree of repeated generation in the resulting images, so we conduct a qualitative comparison to validate the effectiveness of AccDiffusion in eliminating repeated generation.


\begin{table}[!tb]
  \caption{Comparison of quantitative metrics between different training-free image generation extrapolation methods. We use \textbf{bold} to emphasize the best result and \underline{underline} to emphasize the second best result. 
  }
  \label{tab:quantitative comparison}
  \setlength\tabcolsep{2pt}
  \centering
  \begin{tabular}{@{}clcccccc@{}}
    \toprule
    Resolusion & Method & $\text{FID}_r\downarrow $ & $\text{IS}_r\uparrow$  & $\text{FID}_c\downarrow$ & $\text{IS}_c\uparrow$ & CLIP$\uparrow$ & Time\\
    \midrule
    1024 $\times$ 1024 (1$\times$)    & SDXL-DI & 58.49 & 17.39 & 58.08 & 25.38 & 33.07 & <1 min \\
    \midrule
    \multirow{6}{*}{2048 $\times$ 2048 (4$\times$)} & SDXL-DI& 124.40 & 11.05 & 88.33 & 14.64 & 28.11 & 1 min\\
                                                    & Attn-SF & 124.15 & 11.15 & 88.59 & 14.81 & 28.12 & 1 min \\
                                                    & MultiDiffusion & 81.46 & 12.43 & 44.80 & 20.99 & 31.82 & 2 min\\
                                                    & ScaleCrafter & 99.47 & 12.52 & 74.64 & 15.42 & 28.82 & 1 min\\ 
                                                    & DemoFusion & \underline{60.46} & \underline{16.45} & \underline{38.55} & \underline{24.17} & \underline{32.21} & 3 min\\ 
                                                    & AccDiffusion & \textbf{59.63} & \textbf{16.48} & \textbf{38.36} & \textbf{24.62} & \textbf{32.79} & 3 min\\
    \midrule
    \multirow{5}{*}{3072 $\times$ 3072 (9$\times$)}  & SDXL-DI & 170.61 & 7.83 & 112.51 & 12.59 & 24.53 & 3 min\\
                                                    & Attn-SF & 170.62 & 7.93 & 112.46 & 12.52 & 24.56 & 3 min\\
                                                    & MultiDiffusion & 101.11 & 8.83 & 51.95 & 17.74 & 29.49 & 6 min\\
                                                    & ScaleCrafter  & 131.42 & 9.62 & 105.79 & 11.91 & 27.22 & 7 min\\  
                                                    & DemoFusion & \underline{62.43} & \underline{16.41} & \underline{47.45} &  \underline{20.42} & \underline{32.25} & 11 min\\ 
                                                    & AccDiffusion & \textbf{61.40} & \textbf{17.02} & \textbf{46.46} & \textbf{20.77} & \textbf{32.82} & 11 min\\
    \midrule
    \multirow{6}{*}{4096 $\times$ 4096 (16$\times$)} & SDXL-DI & 202.93 & 6.13 & 119.54 & 11.32 & 23.06 & 9 min\\
                                                    & Attn-SF & 203.08 & 6.26 & 119.68 & 11.66 & 23.10 & 9 min\\
                                                    & MultiDiffusion & 131.39 &  6.56 &  61.45 & 13.75 & 26.97 & 10 min\\
                                                    & ScaleCrafter & 139.18 &  9.35 & 116.90 & 9.85 & 26.50 & 20 min\\     
                                                    & DemoFusion & \underline{65.97} & \underline{15.67} & \underline{59.94} & \underline{16.60} & \underline{33.21} & 25 min\\ 
                                                    & AccDiffusion & \textbf{63.89} & \textbf{16.05} & \textbf{58.51} & \textbf{16.72} & \textbf{33.79} & 26 min\\                                       
  \bottomrule
  \end{tabular}
\end{table}

\begin{figure}[!t]
    \centering
    \includegraphics[width=\linewidth]{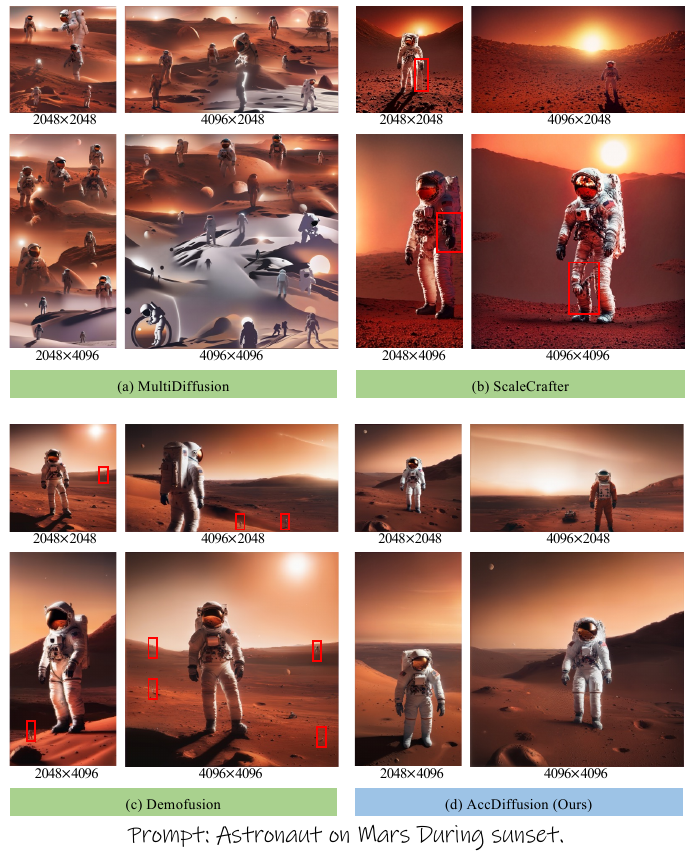}
    \caption{Qualitative comparison of our AccDiffusion with existing training-free image generation extrapolation methods~\cite{bar2023multidiffusion,he2023scalecrafter,du2023demofusion}. We draw a red box upon the generated images to highlight the repeated objects. Best viewed zoomed in.
    }
    \label{fig:Qualitative Comparison}
\end{figure}

\subsection{Qualitative Comparison}
In Fig.\,\ref{fig:Qualitative Comparison}, AccDiffusion is compared with other training-free text-to-image generation extrapolation methods, such as MultiDiffusion~\cite{bar2023multidiffusion}, ScaleCrafter~\cite{he2023scalecrafter}, and DemoFusion~\cite{du2023demofusion}. We provide more results in Appendix\,\ref{app:more visualization} and Appendix\,\ref{app:more results}.
MultiDiffusion can generate seamless images but suffers severe repeated and distorted generation. 
ScaleCrafter, while avoiding the repetition of astronauts, suffers from structural distortions as highlighted in the red box, resulting in local repetition and a lack of coherence. 
DemoFusion tends to generate small, repeated astronauts, with the frequency of repetition escalating with image resolution, thereby significantly degrading image quality. Conversely, AccDiffusion demonstrates superior performance in generating high-resolution images without such repetitions. 
%
%
As Attn-SF~\cite{jin2023logn} and SDXL-DI~\cite{podell2023sdxl} cannot alleviate the repetition issue, their qualitative results are not compared here.

\subsection{Ablation Study}
\label{ablaiton study}
%
In this section, we first perform ablation studies on the two core modules proposed in this paper, and then discuss the settings of the threshold for the binary mask in Eq.\,(\ref{eq:highly responsive regions}) and the threshold $c$ for deciding patch-content-aware prompt in Eq.\,(\ref{eq:patch-content-aware prompt}). All experiments are carried out at a resolution of $4096^2$ ($16 \times$). 
Considering the fact that existing quantitative metrics are unable to accurately reflect the extent of object repetition, we choose to provide visualizations to demonstrate the effectiveness of our core modules in preventing repeated generation.

\subsubsection{Ablations on Core Modules.} 
\begin{figure}[!t]
    \centering
    \includegraphics[width=\linewidth]{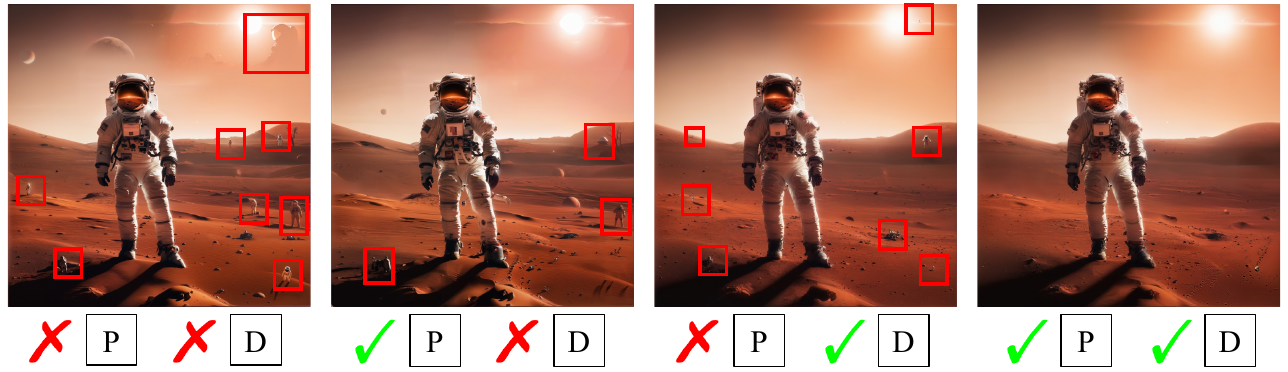}
    \caption{Ablations of Patch-content-aware prompts (\boxed{\text{P}}) and Dilated sampling with window interaction (\boxed{\text{D}}). The ``{\color{red}\XSolidBrush}''/``{\color{green}\Checkmark}''  denotes removing/preserving the component. The repeated objects are highlighted by a red box. Best viewed zoomed in.}
    \label{fig:acc_ablation}
\end{figure}

As illustrated in Fig.\,\ref{fig:acc_ablation}, the absence of any module leads to a decline in generation quality. Without patch-content-aware prompts, the resulting image contains numerous repeated small objects, emphasizing the importance of patch-content-aware prompts in preventing the generation of repetitive elements. 
Conversely, without our window interaction in dilated sampling, the generated small object becomes unrelated to the image, indicating that dilated sampling with window interaction enhances the image's semantic consistency and suppresses repetition.
The maximum number of repeated objects is produced when both modules are removed, while employing both modules simultaneously generates an image free of repetitions. This implies that the two modules work together to effectively alleviate repetitive objects.

\begin{table}[!t]
    \caption{Statistics of cross-attention maps $\mathcal{M}$ using prompt $y$ = ``Astronaut on mars during sunset.'' as an example. Each word $\{y_j\}^6_{j=1}$ has a cross-attention map ${\{\mathcal{M}}_{:,j}\}^6_{j=1}$.}
    \centering
    \setlength\tabcolsep{3pt}
    \begin{tabular}{@{}ccccccc@{}}
        \toprule
        \multirow{2}{*}{Statistics}  & ``Astronaut'' & ``on''    & ``mars'' & ``during'' & ``sunset'' & ``.''   \\
                               &   $(j=1)$     &  $(j=2)$  &  $(j=3)$ &  $(j=4)$   &  $(j=5)$   & $(j=6)$ \\
         \midrule
         $ \text{Min}({\mathcal{M}}_{:,j})$ & 0.1274 & 0.0597 & 0.2039 & 0.0457 & 0.0921 & 0.0335 \\
         $ \text{Mean}({\mathcal{M}}_{:,j})$ & 0.1499 & 0.0676 & 0.2533 & 0.0521 & 0.1189 & 0.0386 \\
         $\text{Max}({\mathcal{M}}_{:,j})$ & 0.2096 & 0.0779 & 0.2979 & 0.0585 & 0.1499 & 0.0419 \\
         \bottomrule
    \end{tabular}
    \label{tab:ablation for highly responsive threshold}
\end{table}

\subsubsection{Ablations on Hyper-Parameters.}
%

As depicted in Table\,\ref{tab:ablation for highly responsive threshold}, there is a significant variation in the range of different cross-attention maps $\mathcal{M}_j$. When using a fixed threshold for these maps, two scenarios may occur. If the threshold is too high, some words will not have highly responsive regions in their corresponding attention maps, resulting in their absence from the patch-content-aware prompt. Conversely, if the threshold is too low, the entire attention map consists of highly responsive regions, causing those words to consistently appear in the patch-content-aware prompt. By considering the average $\overline{\mathcal{M}}_{:,j}$, we can ensure that each word has suitable highly responsive regions, as demonstrated in Fig.\,\ref{fig:attention_map}(b). 

\begin{figure}[!t]
    \centering
    \includegraphics[width=\linewidth]{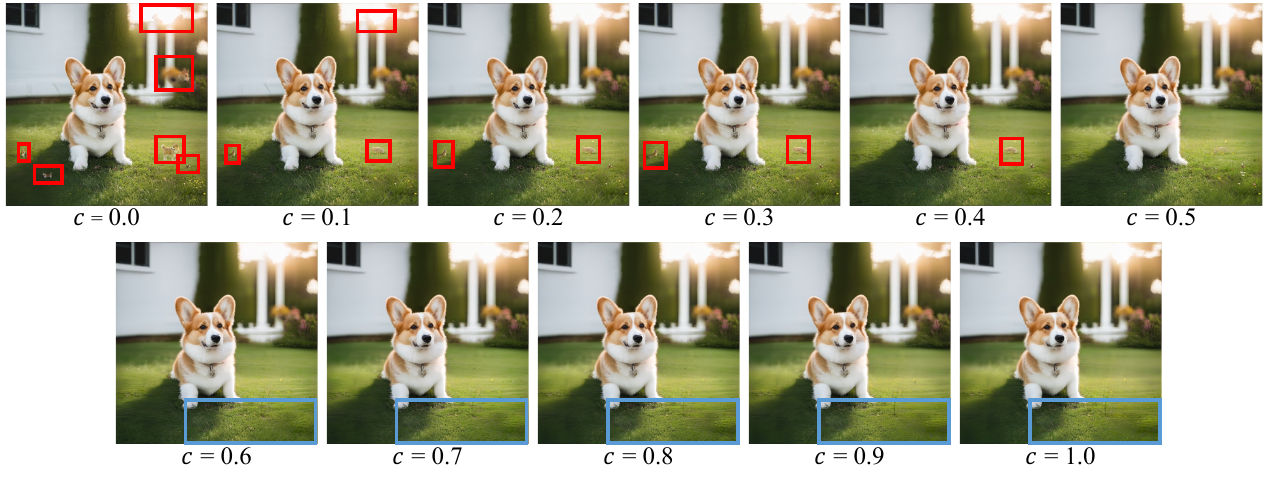}
    \caption{Visual results of different threshold $c$, prompted by ``A cute corgi on the lawn.'' The repeated objects are highlighted with a red box and the detail degradation is stressed with a blue box. Best viewed zoomed in.}
    \label{fig:different_c}
\end{figure}

Recall in Eq.\,(\ref{eq:patch-content-aware prompt}), the $c$ determines whether the proportion of a highly responsive region for a word $y_j$ surpasses the threshold required for inclusion in the prompts of patch $z_t^i$. 
A very small value of $c$ leads to more words being included in the patch prompt, potentially causing object repetition. 
Conversely, a very large value of $c$ simplifies the patch prompt, which may lead to degradation of details.
Our analysis is demonstrated in Fig.\,\ref{fig:different_c}. 
It should be noted that this is a user-specific hyper-parameter, adjustable to suit different application scenarios.

\section{Limitations and Future work}

AccDiffusion is limited in: (1) As it follows the DemoFusion pipeline, similar drawbacks arise such as inference latency from progressive upscaling and overlapped patch-wise denoising. (2) AccDiffusion focuses on image generation extrapolation, meaning the fidelity of high-resolution images depends on the pre-trained diffusion model. (3) Relying on LDMs' prior knowledge of cropped images, it may produce local irrational content in sharp close-up image generation.

Future studies could explore the possibility of developing non-overlapped patch-wise denoising techniques for efficiently generating high-resolution images.

\section{Conclusion}
In this paper, we propose AccDiffusion to address the object-repeated generation issue in higher-resolution image generation without training.
AccDiffusion first introduces patch-content-aware prompts, which makes the patch-wise denoising more accurate and can avoid repeated generation from the root. 
And then we further propose dilated sampling with window interaction to enhance the global consistency during higher-resolution image generation.
Extensive experiments, including qualitative and quantitative results, show that AccDiffusion can successfully conduct higher-resolution image generation without object repetition.

\par\vfill\par

\section*{Acknowledgements}
This work was supported by National Science and Technology Major Project (No. 2022ZD0118202), the National Science Fund for Distinguished Young Scholars (No.62025603), the National Natural Science Foundation of China (No. U21B2037, No. U22B2051, No. U23A20383, No. 62176222, No. 62176223, No. 62176226, No. 62072386, No. 62072387, No. 62072389, No. 62002305 and No. 62272401), and the Natural Science Foundation of Fujian Province of China (No.2022J06001).


%
%
\bibliographystyle{splncs04}
\bibliography{main}

\clearpage

\begin{center}
\textbf{\Large Appendix \\}
\end{center}
\appendix

\begin{figure}[h]
    \centering
    \includegraphics[width=\linewidth]{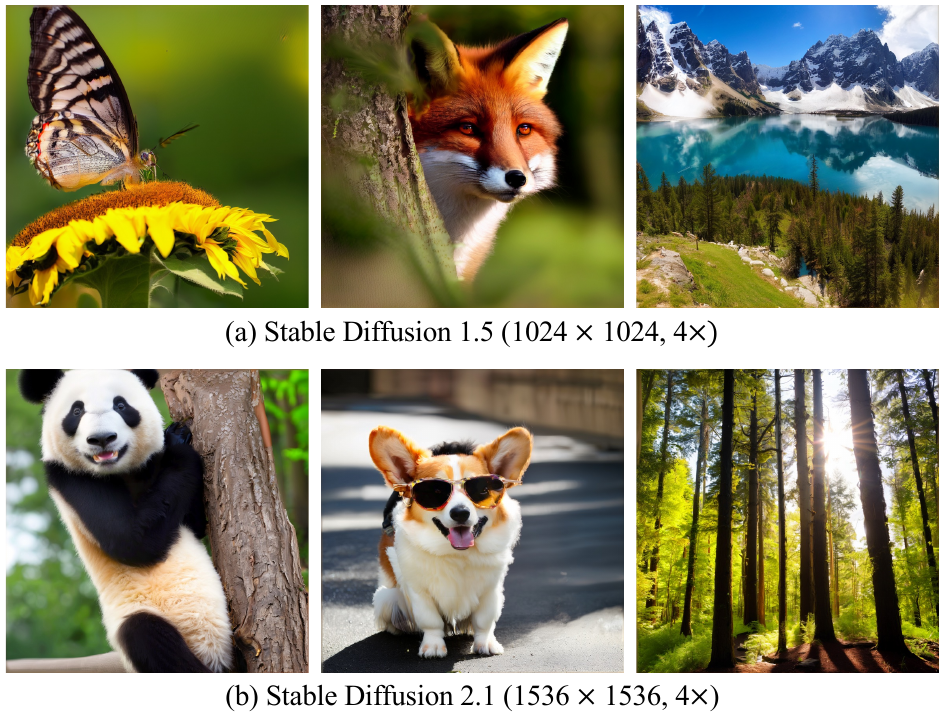}
    \caption{Results of AccDiffusion on other stable diffusion variants: (a) Stable diffusion 1.5 (default resolution of $512^2$) and (b) Stable diffusion 2.1 (default resolution of $768^2$). All images are generated at $4 \times$ resolution. Prompts are provided in Sec.\,\ref{sec:prompt_zoo}.
    }
    \label{fig:variants}
    \vspace{-2em}
\end{figure}

\section{More Stable Diffusion Variants}\label{more variants}
We apply AccDiffusion on other LDMs, specifically Stable Diffusion 1.5 (SD 1.5)~\cite{stable-diffusion-1.5} and Stable Diffusion 2.1~\cite{stable-diffusion-2-1} (SD 2.1).
As shown in Fig.\,\ref{fig:variants}, AccDiffusion successfully generates higher-resolution images without repetition.
It is important to note that the results of AccDiffusion depend on the prior knowledge of LDMs, and the performance of SD 1.5 and SD 2.1 is inferior to SDXL~\cite{podell2023sdxl}.
Therefore, the fidelity of their results  are less astonishing than those on SDXL.

\section{More Visualization}\label{app:more visualization}
\begin{figure}[t]
    \centering
    \includegraphics[width=\linewidth]{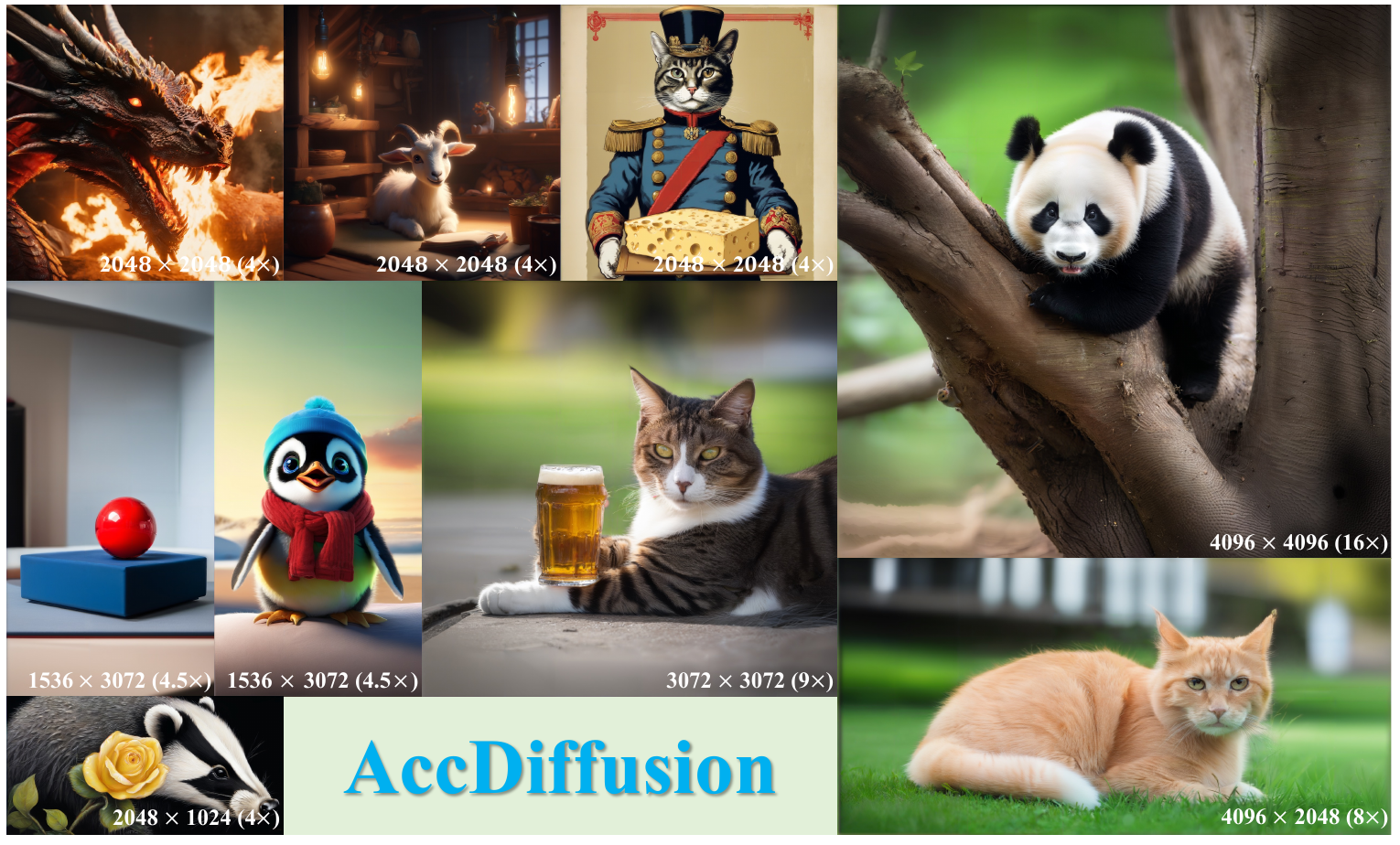}
    \caption{More higher-resolution results of AccDiffusion on SDXL (default resolution of $1024^2$). Best viewed with zooming in.
    }

    \label{fig:more_visualization}
\end{figure}

We provide more results of AccDiffusion on SDXL. As shown in Fig.\,\ref{fig:more_visualization}, our AccDiffusion can generate various higher-resolution images without object repetition. Prompts are provided in Sec.\,\ref{sec:prompt_zoo}.

\section{Default Setting of DemoFusion}\label{app:default setting}
To ensure a fair comparison with DemoFusion~\cite{du2023demofusion}, we conduct our experiments using its default settings listed in Table\,\ref{tab:setting}. For a more comprehensive understanding of DemoFusion, please refer to the original paper~\cite{du2023demofusion}.

\begin{table}[!h]
    \caption{The default setting of DemoFusion~\cite{du2023demofusion}.}
    \centering
    \setlength\tabcolsep{2pt}
    \resizebox{.3\columnwidth}{!}{
    \begin{tabular}{ccc}
    \toprule
    Parameters& Explanation & Values \\
    \midrule
    $T$  & DDIM Steps &  50 \\
    $s$  & Guidance Scale & 7.5 \\
    $h$  & Latent Height & 128 \\
    $w$  & Latent Width & 128 \\
    $d_h$ & Height Stride  & $\frac{h}{2}$ \\
    $d_w$ & Width Stride  & $\frac{w}{2}$ \\
    $\alpha_1$  & Scale factor 1 & 3 \\
    $\alpha_2$  & Scale factor 2 & 1 \\
    $\alpha_3$  & Scale factor 3 & 1 \\
    \bottomrule
    \end{tabular}
    }
    \label{tab:setting}
\end{table}

\section{More Qualitative Comparison Results}\label{app:more results}
\begin{figure}[!t]
    \centering
    \includegraphics[width=\linewidth]{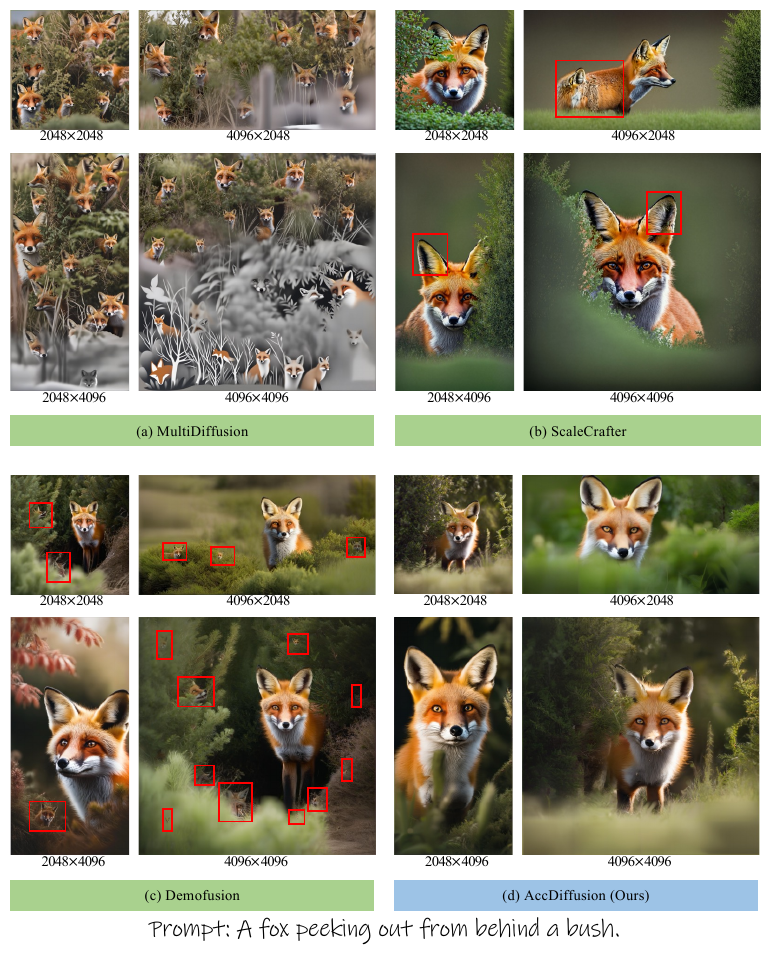}
    \caption{Qualitative comparison of our AccDiffusion with existing training-free image generation extrapolation methods~\cite{bar2023multidiffusion,he2023scalecrafter,du2023demofusion}. We draw a red box upon the generated images to highlight the repeated objects. Best viewed zoomed in.
    }
    \label{fig:Qualitative Comparison fox}
\end{figure}

\begin{figure}[!t]
    \centering
    \includegraphics[width=\linewidth]{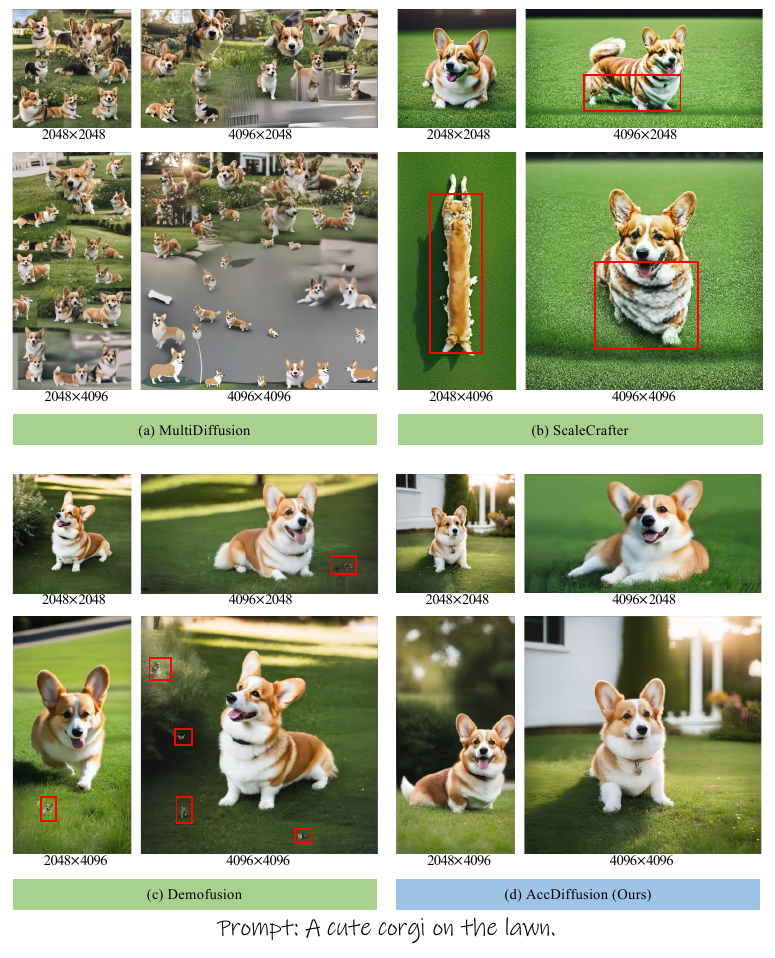}
    \caption{Qualitative comparison of our AccDiffusion with existing training-free image generation extrapolation methods~\cite{bar2023multidiffusion,he2023scalecrafter,du2023demofusion}. We draw a red box upon the generated images to highlight the repeated objects. Best viewed zoomed in.
    }
    \label{fig:Qualitative Comparison dog}
\end{figure}

We provide more qualitative comparison results in Fig.\,\ref{fig:Qualitative Comparison fox} and Fig.\,\ref{fig:Qualitative Comparison dog}. More qualitative results provide stronger evidence that our method can generate high-resolution images without repetition.

\section{Details on any aspect ratio generation.} 

First,  we initialize a latent noise with the expected ratio and set the longer side to training resolution (\emph{e.g.}, $1024\times512$ for 2:1).  Then we use the same pipeline as the 1:1 aspect ratio to progressively generate higher-resolution images, as shifted window sampling and dilated sampling are compatible with any aspect ratio. More details can be found in DemoFusion~\cite{du2023demofusion}.

\begin{figure}[!t]
    \centering
    \includegraphics[width=\linewidth]{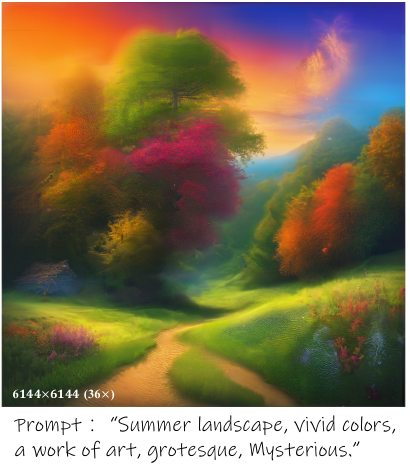}
    \caption{Failure case of indefinite extrapolation.}
    \label{fig:Failure case}
\end{figure}

\section{Indefinite extrapolation.} 
Following the recent works, we provide results in main paper within 4K for comparisons.
Ideally, both AccDiffusion and patch-wise methods can extrapolate indefinitely. 
However, we find that AccDiffusion faces detail degradation when the resolution is beyond 6K (36$\times$), as shown in Fig.\,\ref{fig:Failure case}.

\section{Pseudo Code of AccDiffusion}
AccDiffusion follows the pipeline of DemoFusion~\cite{du2023demofusion} and uses the patch-content-aware prompts during the progress of higher-resolution image generation. Additionally, AccDiffusion enhances dilated sampling with window interaction. 
Algorithm\,\ref{alg:accdiffusion} illustrates the process of higher-resolution generation using AccDiffusion.
We use red color to highlight two core modules proposed by AccDiffusion.

\section{Prompts Used in Supplement Material}
\label{sec:prompt_zoo}
\subsubsection{Fig.\,\ref{fig:variants}:}
\begin{enumerate}
    \item A butterfly landing on a sunflower.
    \item A fox peeking out from behind a bush.
    \item A picturesque mountain scene with a clear lake reflecting the surrounding peaks.
    \item A cute panda on a tree trunk. 
    \item A corgi wearing cool sunglasses. 
    \item Primitive forest, towering trees, sunlight falling, vivid colors.
\end{enumerate}
\subsubsection{\cref{fig:more_visualization}:}
\begin{enumerate}
    \item A close-up of a fire spitting dragon, cinematic shot.
    \item Cute adorable little goat, unreal engine, cozy interior lighting, art station, detailed’ digital painting, cinematic, octane rendering.
    \item A propaganda poster depicting a cat dressed as french emperor napoleon holding a piece of cheese.
    \item A cute panda on a tree trunk.  
    \item a photograph of a red ball on a blue cube.
    \item a baby penguin wearing a blue hat, red gloves, green shirt, and yellow pants.
    \item a cat drinking a pint of beer.
    \item A young badger delicately sniffing a yellow rose, richly textured oil painting.
    \item A cute cat on the lawn. 
\end{enumerate}

\begin{algorithm*}[h]
    \caption{The process of higher-resolution generation using AccDiffusion}\label{alg:accdiffusion}
    \begin{algorithmic}[1]
    \Statex \textbf{Input}:\quad $h'$, $w'$  \Comment{Latent Size of Desired Image }
    \Statex \quad\quad\quad\quad $\mathcal{E}_{\theta}$, $h$, $w$ \Comment{Pre-trained Stable diffusion and Pre-trained Latent Size}
    \Statex \quad\quad\quad\quad $y$, $\mathcal{D}$  \Comment{Prompt and Decoder}
    \Statex \quad\quad\quad\quad $\eta_1$, $\eta_2$  \Comment{Decreasing From $1$ to $0$ Using a Cosine Schedule}
    \State \texttt{\#\#\#\#\#\#\#\#\#\#\#\#\#\# Phase $1$: Low resolution image generation 
                   \#\#\#\#\#\#\#\#\#\#\#\#\#\#}
    \State $\mathbf{z}_T \sim \mathcal{N}(0, I)$ \Comment{Random Initialization}
    \For{$t = T$ to $1$}
        \State     $\mathbf{z}_{t-1} = \sqrt{\frac{\alpha_{t-1}}{\alpha_t}}\mathbf{z}_t + \Bigg(\sqrt{\frac{1}{\alpha_{t-1}} - 1} - \sqrt{\frac{1}{\alpha_t} - 1} \Bigg)\cdot \varepsilon_{\theta}\big(\mathbf{z}_t,t,\tau_{\theta}(y)\big).$ \\ \Comment{Denoising with Image-content-aware Prompt and Save Cross-Attenion Map $\mathcal{M}$}
    \EndFor
    \State $\mathcal{Z}_0 = \mathbf{z}_0$

    \State $S$ = $\frac{h'}{h} \times \frac{w'}{w}$  \Comment{Progressive Upscaling Times}
    \State \texttt{\#\#\#\#\#\#\#\#\#\#\#\# Phase $2$: Higher-resolution image generation 
                   \#\#\#\#\#\#\#\#\#\#\#\#\#}
    \For{$s = 2$ to $S$} \Comment{Progressive Upscaling}
        \State $\mathcal{Z}_0 =  inter(\mathcal{Z}_0, (h\times s, w\times s))$ \Comment{Interpolation Upsampling}
        \For{$t = 1$ to $T$}
        \State $\mathcal{Z}'_{t} = \sqrt{\bar{\alpha}_t}\mathcal{Z}_0 + \sqrt{1-\bar{\alpha}_t}\varepsilon,\quad \varepsilon\sim \mathcal{N}(0,\mathbf{I})$\\
                \Comment{Getting Noise-inversed Representations }
        \EndFor
        \State $\mathcal{Z}_T = \mathcal{Z}'_T$
        \For{$t = T$ to $1$}
            \State ${\hat{\mathcal{Z}}}_t = \eta_1 \times \mathcal{Z}'_t  + (1-\eta_1) \times \mathcal{Z}_t$ \Comment{Skip Residual}
            \State $\{\mathbf{z}^i_t\}^{P_1}_{i=1}$ = $\text{Sampling}_1(\hat{\mathcal{Z}}_t) $
                        \Comment{Shift Window Sampling From MultiDiffusion}
            \State $\mathcal{M} \rightarrow \{\gamma^i\}^{P_1}_{i=1}$ \Comment{Calculating {\color{red}Patch-Content-Aware Prompt}}
            \State $\{\mathcal{D}^i_t\}^{P_2}_{i=1}$ = $\text{Sampling}_2(\hat{\mathcal{Z}}_t) $
                        \Comment{Dilated  Sampling From DemoFusion}
            \For{$\mathbf{z}^i_t$ in $\{\mathbf{z}^i_t\}^{P_1}_{i=1}$}\\
                 \quad\quad\quad\quad$\mathbf{z}^i_{t-1} = \sqrt{\frac{\alpha_{t-1}}{\alpha_t}}\mathbf{z}^i_t + \Bigg(\sqrt{\frac{1}{\alpha_{t-1}} - 1} - \sqrt{\frac{1}{\alpha_t} - 1} \Bigg)\cdot \varepsilon_{\theta}\big(\mathbf{z}^i_t,t,\tau_{\theta}(\gamma^i)\big).$ 
                 \\  \Comment{Denoising with {\color{red}Patch-Content-Aware Prompt}}
            \EndFor
            \For{$\mathcal{D}^i_t$ in $\{\mathcal{D}^i_t\}^{P_2}_{i=1}$}\\
                \quad\quad\quad\quad ${\mathcal{D}_t}^{k,h,w} = {\mathcal{D}_t}^{f_t^{h,w}(k),h,w}$\Comment{{\color{red}Window Interaction with Bijective Function}}\\ 
                \quad\quad\quad\quad$\mathcal{D}^i_{t-1} = \sqrt{\frac{\alpha_{t-1}} {\alpha_t}}\mathcal{D}^i_t + \Bigg(\sqrt{\frac{1}{\alpha_{t-1}} - 1} - \sqrt{\frac{1}{\alpha_t} - 1} \Bigg)\cdot \varepsilon_{\theta}\big(\mathcal{D}^i_t,t,\tau_{\theta}(y)\big)$ \\ \Comment{Denoising with Image-Content-Aware Prompt}\\
                \quad\quad\quad\quad    
                    ${\mathcal{D}_{t-1}}^{k,h,w} = {\mathcal{D}_{t-1}}^{{(f^{h,w}_t)}^{-1}(k),h,w}$\Comment{{\color{red} Recover}}
                \\  
            \EndFor
            \State  ${\mathcal{Z}}_{t-1} = \eta_2\times\text{Fuse}( \{\mathcal{D}^i_t\}^{P_2}_{i=1}) + (1 - \eta_2) \times \text{Fuse}(\{\mathbf{z}^i_t\}^{P_1}_{i=1})$ \\
            \Comment{Fusing Shift Window Sampling Patches and Dilated  Sampling Patches}
        \EndFor
    \EndFor
    \Statex \textbf{Output}: $\mathbf{x}_0=\mathcal{D}(\mathcal{Z}_0)$ 
    \Comment{Decoding to Image}
    \end{algorithmic}

\end{algorithm*}

\end{document}